%% file: main.tex
\documentclass[letterpaper, 10 pt, conference]{ieeeconf}
\IEEEoverridecommandlockouts
\usepackage{cite}
\usepackage{amsmath,amssymb,amsfonts}
\usepackage{graphicx}
\usepackage{booktabs}
\usepackage{multirow}
\usepackage{xcolor}
\usepackage{url}
\usepackage[hidelinks]{hyperref}

\newcommand{\dataset}{\textsc{AquaAct}}

\title{\LARGE \bf
\dataset{}: A Joystick-Level Underwater Teleoperation Dataset and Closed-Loop Benchmark
}

\author{Zhenqi Wu$^{1}$, Trung Tien Dong$^{1}$, Aditya Penumarti$^{2}$, Mohamed A. Fouad$^{1}$,\\
Menna A. Fouad$^{1}$, Mingyang Mao$^{1}$, Jane Jaejeong Shin$^{3}$, and Xiaomin Lin$^{1}$%
\thanks{$^{1}$University of South Florida, Tampa, FL, USA.}%
\thanks{$^{2}$University of Florida, Gainesville, FL, USA.}%
\thanks{$^{3}$Seoul National University, Seoul, Republic of Korea.}%
}

\begin{document}
\bstctlcite{IEEEexample:BSTcontrol}
\maketitle
\thispagestyle{empty}
\pagestyle{empty}

\input{sections/00_abstract}

\input{sections/01_introduction}
\input{sections/02_related_work}
\input{sections/03a_simulation}
\input{sections/03_dataset}
\input{sections/05_exp1_sim_value}
\input{sections/06_exp2_benchmark}
\input{sections/08_conclusion}

\bibliographystyle{IEEEtran}
\bibliography{refs}

\end{document}

%% file: sections/00_abstract.tex
\begin{abstract}
Underwater robots support a growing range of work, from inspecting offshore structures and
pipelines to surveying reefs and accompanying divers. Most of these tasks, however, are
still flown by hand by a pilot, which raises the question of whether they can be carried out
autonomously instead. On land,
vision-language-action (VLA) models already let robots carry out tasks autonomously from language instructions, which makes them a natural candidate for taking
the pilot's place.  Underwater, however, there is little data to
train such models. We take inspiration from how these tasks are flown today: the pilot watches the camera image
and steers with a joystick, whose commands the flight controller turns into motion. We
therefore record the pilots and introduce \dataset{}, to our knowledge the first publicly
available language-annotated underwater dataset pairing camera images with human joystick
commands. It comprises 2.0 hours of real-world and 17.9 hours of simulated demonstrations,
both flown by human pilots, together with the vehicle's inertial measurements. Experiments with \dataset{} show that its simulated demonstrations transfer to real data and
greatly reduce the real demonstrations a policy needs. A closed-loop benchmark of seven policies in simulation, from imitation-learning baselines
to VLA models, each trained with its default recipe, shows that these tasks remain far from
solved: no policy completes more than a fifth of the episodes. Pretraining on land data gives
large VLA models no advantage over small policies trained from scratch, a gap that
land-trained models have yet to close underwater.
\end{abstract}

%% file: sections/01_introduction.tex
\section{Introduction}
\label{sec:intro}

Underwater inspection, monitoring, and surveying are essential for
maintaining marine infrastructure and studying ocean environments:
subsea pipelines, offshore wind foundations, and aquaculture facilities
require regular inspections, while reef monitoring, shipwreck
exploration, and diver assistance demand operations in diverse and
often challenging environments~\cite{auv_inspection_review,offshore_inspection_review,netpen_visual_servo,oystersim,dream_uwmonitor,caddy}.
These environments often exceed the depth, bottom time, and risk
that human divers can safely accept~\cite{osha_diving}, so underwater
robots increasingly take over this work~\cite{petillot2019}.
However, most of these operations still rely on remotely operated
vehicles (ROVs) flown by human pilots through onboard camera feedback,
so pilot workload limits the scale and duration of these
missions~\cite{offshore_inspection_review,sharc,subsea_hmi_review}.

Reducing this dependence on human pilots requires autonomous systems
that interpret underwater scenes, follow high-level instructions, and
act accordingly.
Conventional underwater autonomy builds a perception--planning--control
pipeline for each task and sensor
setup~\cite{hover2012hull,ridao_iauv,netpen_visual_servo}, which must be
adapted for every new environment or task~\cite{uw_full_autonomy_review}.
Vision-language-action (VLA) models offer an alternative: they learn to
predict robot actions from visual observations and language
instructions~\cite{rt2,openvla,octo,pi0,pi05,gr00t,smolvla} and,
trained on large demonstration datasets such as
Open X-Embodiment~\cite{oxe} and BridgeData V2~\cite{bridgev2}, have
shown that skills transfer across tasks and
robots~\cite{octo,robot_vla_survey}.

\begin{figure}[t]
  \centering
  \vspace{0.1in}
  \includegraphics[width=\linewidth]{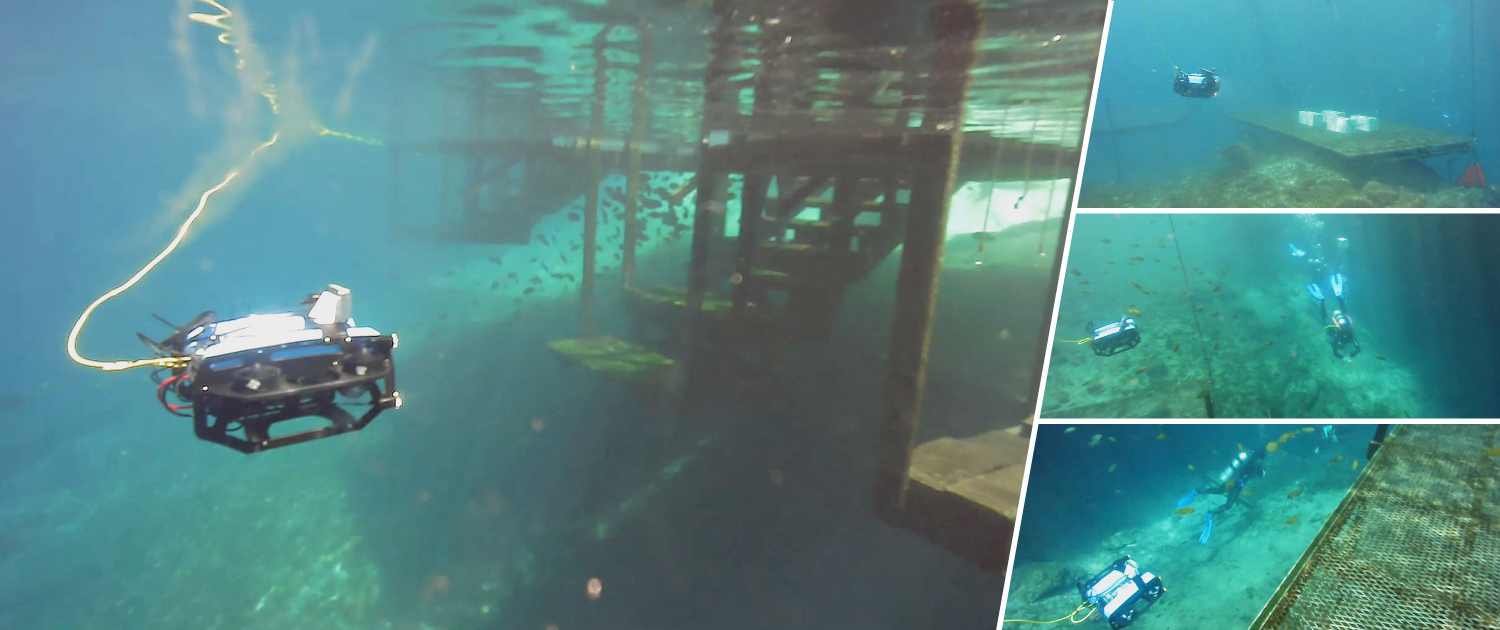}
  \caption{The BlueROV2 Heavy at Blue Grotto, Florida: scanning the submerged deck (left),
  scanning the concrete block (top right) and following divers (middle and bottom right).}
  \vspace{-0.15in}
  \label{fig:teaser}
\end{figure}

Despite these advances, underwater robotics lacks comparable
demonstration data for training general-purpose policies.
Collecting real underwater demonstrations requires deploying a vehicle
and a skilled pilot, making large-scale data acquisition costly and
difficult.
For an underwater vehicle, the natural action for such data is the
pilot's joystick command.
It is what the pilot actually issues in response to the onboard
camera view.
It also leaves stabilization to the vehicle's flight controller,
which holds attitude and depth in a control loop far faster than a
policy can run.
The policy then only decides where the vehicle should go, much as
VLA models on land output end-effector targets and leave joint
control to the robot's own controllers.
Existing underwater learning datasets provide only parts of this.
The closest vehicle-level effort relies entirely on simulation,
with demonstrations generated by scripted experts and actions given
as thruster commands~\cite{usim_u0}.
Real-world imitation learning with recorded operator commands has
primarily focused on manipulation tasks in controlled pools or
tanks~\cite{uw_aquabot,uw_biaqua,uw_uloha}, while underwater
navigation datasets have relied on retrospectively annotated actions
rather than the actual commands issued by human
pilots~\cite{uw_nav2goal,uw_uivnav}.
Consequently, to our knowledge, no publicly available
language-annotated dataset provides synchronized onboard observations
and pilot-issued joystick commands for learning real-world underwater
vehicle action.

We therefore collect such data by recording the pilots themselves,
on the real vehicle and in simulation.
Recording the onboard camera view together with the pilot's joystick
commands yields paired observations and actions directly, without
manual action labeling, and we annotate each task with a language
instruction that states its goal, such as ``Scan the deck from left
to right'' or ``Follow the pipeline''.
Because the control chain, from the joystick through the
ground-control software to the flight-controller firmware, can be
reproduced in simulation, pilots can record further demonstrations
there, at sites a real vehicle reaches only rarely.

Building on this approach, we introduce \dataset{}, comprising
2.0 hours of real-world underwater demonstrations
(Fig.~\ref{fig:teaser}) and 17.9 hours of simulated
demonstrations (Fig.~\ref{fig:system}), with the simulator
calibrated against real-world runs.
This combination allows us to investigate whether pretraining
on underwater demonstrations collected from our simulation pipeline
improves learning from limited real-world data.
We further evaluate how existing general-purpose robot
policies perform when learning to control underwater vehicles
through joystick commands: we train seven policies, from
imitation-learning baselines to VLA models, on the simulated part
with their default recipes and run each for 50 closed-loop episodes
in the same simulation pipeline, where a program scores the success
and completion of every episode from the ground-truth state and a
human reviewer checks each score.
Our contributions are:

\begin{itemize}
    \item \textbf{Underwater Dataset:}
    We introduce \dataset{}, a language-annotated dataset
    of real and simulated ROV demonstrations with synchronized
    visual, inertial, and joystick data (Sec.~\ref{sec:dataset}).

    \item \textbf{Simulation Data Collection:}
    We develop a real-world-calibrated, firmware-in-the-loop
    simulation pipeline for collecting human demonstrations through
    the same control chain as the real ROV
    (Sec.~\ref{sec:sim}, \ref{sec:fidelity}).

    \item \textbf{Sim-to-Real Evaluation:}
    We show that pretraining using data from our simulation pipeline reduces
    real-world data requirements for policy learning
    (Sec.~\ref{sec:exp_real}).

    \item \textbf{Policy Benchmark:}
    We benchmark seven policies, from imitation-learning baselines
    to VLA models, in closed-loop simulation and analyze where they
    fail, revealing a clear gap in transferring land-trained models
    underwater (Sec.~\ref{sec:exp_sim}).
\end{itemize}

%% file: sections/02_related_work.tex
\section{Related Work}
\label{sec:related}

\begin{figure*}[t]
  \centering
  \includegraphics[width=\textwidth]{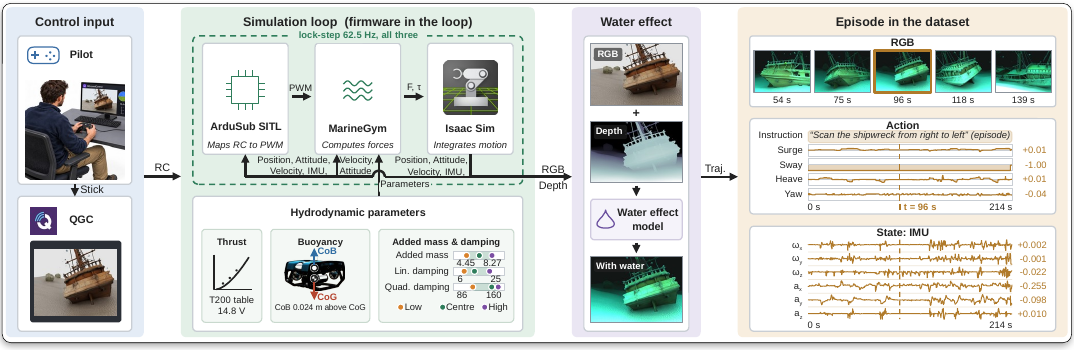}
  \vspace{-0.2in}
  \caption{Firmware-in-the-loop simulation pipeline: the pilot's joystick commands pass through
  QGroundControl and the ArduSub firmware to the simulated vehicle, and the recorder stores the
  camera, inertial and joystick streams (IMU traces shown with gravity removed). Traj.: trajectories.}
  \vspace{-0.2in}
  \label{fig:system}
\end{figure*}

\subsection{Classical Underwater Autonomy}
Autonomous underwater inspection has traditionally been built as a modular
perception--planning--control pipeline engineered around one task, such as ship-hull
inspection~\cite{hover2012hull}, subsea cable tracking~\cite{xiang2016cable},
net-pen inspection~\cite{netpen_visual_servo} and intervention~\cite{ridao_iauv}. Learning has
entered these stacks mainly as better components, such as detectors or
controllers~\cite{uw_ai_review}. These systems are reliable within
their design envelope, but each new task or site calls for new modules and new
tuning~\cite{uw_full_autonomy_review}.

\subsection{Learning from Demonstrations and VLA Models}
Learning from demonstrations removes the need to engineer a pipeline for each task: a policy
maps observations to actions directly, and a new task calls for new data rather than new
modules. Imitation-learning policies such as action-chunking transformers (ACT)~\cite{act}
and diffusion policies~\cite{diffusionpolicy} map images and proprioception to short action
sequences and learn fine-grained skills from a modest number of demonstrations, but each is
typically trained for a narrow task family. Generality has come from data: large corpora such
as BridgeData~V2~\cite{bridgev2} and Open X-Embodiment~\cite{oxe} pool demonstrations across
tasks, scenes and robots, and VLA models build on them by attaching an action output to a
pretrained vision-language backbone, so that one model follows language instructions across
tasks. RT-2~\cite{rt2} emits actions as tokens, OpenVLA~\cite{openvla} and Octo~\cite{octo}
are open, fine-tunable generalist policies, $\pi_0$~\cite{pi0} and $\pi_{0.5}$~\cite{pi05}
add a flow-matching action expert, GR00T~N1~\cite{gr00t} targets humanoids,
SmolVLA~\cite{smolvla} shows that a compact model can be competitive, and
MemoryVLA~\cite{memoryvla} adds a memory of past observations. Vision-language navigation
models~\cite{navid,uninavid,omninav} also follow instructions through space, but they output
waypoints or motion primitives for a lower layer to execute rather than the continuous
commands a pilot issues. All of these models, however, are trained on terrestrial data; none
of their corpora contains underwater demonstrations or uses an operator's joystick as the
action space.

\subsection{Underwater Robot Learning and Simulation}
Underwater, learning from demonstrations is still small in scale. AquaBot~\cite{uw_aquabot}
records a pilot's gamepad commands with video for about a thousand manipulation
demonstrations in a pool, without language, and imitation learning for underwater arms has
likewise been shown in pools and tanks~\cite{uw_biaqua,uw_uloha}, including a VLA evaluated
on real underwater manipulation~\cite{uw_uloha}. For navigation, real-world learning has used
camera footage whose actions were labeled afterwards~\cite{uw_nav2goal,uw_uivnav}, and
Khanmeh et~al.~\cite{uw_rovtether_scidata} released BlueROV2 pilot inputs without camera
images. Simulation is the standard way to scale beyond this. MarineGym~\cite{marinegym} adds
hydrodynamics to GPU physics for reinforcement learning, and USIM~\cite{usim_u0}, the closest
vehicle-level effort toward underwater VLA, generates 20 navigation and manipulation tasks in
simulation with hand-crafted experts that use ground-truth state and fine-tunes U0 from
GR00T~N1.5 to map images, language and sensor readings to per-thruster commands and arm joint
angles, without real-world deployment. Thus real-world work records a human operator's
commands only for manipulation in pools or tanks; no existing dataset pairs open-water
observations with pilot commands for vehicle-level tasks, in reality or in simulation.

Filling this gap in simulation, where demonstrations are cheap, adds a further requirement:
simulated demonstrations help if the simulated vehicle moves like the real
one, which depends on the hydrodynamic model~\cite{fossen} and on the controller in the loop,
and looks like it, which depends on how water attenuates and scatters
light~\cite{akkaynak2018,seathru}.
Taken together, these lines of work point to four needs that guide our collection pipeline
and dataset: (i) real underwater observations paired with the joystick commands a human pilot
actually sent; (ii) simulated demonstrations recorded through the same control chain, from
the pilot's joystick through the ground-control software to the flight-controller firmware;
(iii) a simulator whose dynamics and appearance are calibrated against the real vehicle; and
(iv) an evaluation of how current policies, developed for other action spaces, learn from the
pilot's commands.
Table~\ref{tab:compare} summarizes how \dataset{} differs from the closest datasets.

\begin{table}[t]
  \centering
  \caption{Comparison with related datasets.}
  \label{tab:compare}
  \setlength{\tabcolsep}{3pt}
  \resizebox{\linewidth}{!}{%
  \begin{tabular}{lccccc}
    \toprule
    Dataset & Underwater & Real & Sim & Action space & Lang. \\
    \midrule
    Open X-Emb.~\cite{oxe}   & -- & \checkmark & -- & EE / joint & \checkmark \\
    USIM~\cite{usim_u0}      & \checkmark & -- & \checkmark & Thruster PWM, arm & \checkmark \\
    AquaBot~\cite{uw_aquabot} & \checkmark & \checkmark & -- & Gamepad, arm (pool) & -- \\
    Khanmeh~\cite{uw_rovtether_scidata} & \checkmark & \checkmark & -- & RC, no video & -- \\
    \dataset{} (ours)        & \checkmark & \checkmark & \checkmark & Joystick (RC) & \checkmark \\
    \bottomrule
  \end{tabular}}
  \vspace{-0.2in}
\end{table}

\begin{figure*}[t]
  \centering
  \includegraphics[width=\textwidth]{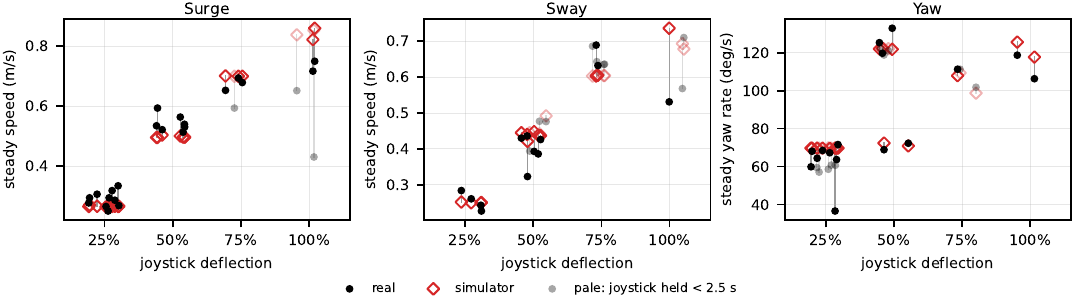}
  \caption{Steady response to held joystick inputs: real vehicle (black) versus simulator
  (red); surge and sway speed from the Doppler log, yaw rate from the IMU. Pale: input held
  for less than 2.5\,s.}
  \label{fig:steady}
  \vspace{-0.1in}
\end{figure*}

%% file: sections/03a_simulation.tex
\section{Platform and Firmware-in-the-Loop Simulation}
\label{sec:platform}

To meet these needs, real and simulated demonstrations pass through the same control chain,
from the pilot's joystick through the ground-control software to the flight-controller
firmware, and the simulator is calibrated against the real vehicle in motion and in
appearance.

\subsection{Platform and Action Interface}
\label{sec:interface}
Real and simulated demonstrations both use a BlueROV2 Heavy, an eight-thruster vehicle whose
flight controller runs ArduSub~\cite{ardusub}. The pilot moves a joystick connected to the
ground-control software, QGroundControl~\cite{qgroundcontrol}, which sends the joystick positions to the flight
controller as RC channel commands; all recordings use the flight controller's depth-hold
mode. We record the four channels the pilot commands, surge, sway, heave and yaw, as pulse
widths in $[1100,1900]\,\mu$s and normalize them as $a=(\mathrm{pwm}-1500)/400\in[-1,1]$;
roll and pitch are held by the flight controller and are not part of the action. The
observation is the forward camera image and a six-dimensional inertial state: body angular
rates and linear acceleration, including gravity, from the IMU. Streams are recorded at
15\,Hz and exported at 10\,Hz after time alignment.

\subsection{Firmware-in-the-Loop Simulation Pipeline}
\label{sec:sim}
Real demonstrations are expensive and limited to the sites a team can reach, so we build a
simulation pipeline that pilots drive through the same joystick, ground-control software and
flight-controller firmware (Fig.~\ref{fig:system}). In this pipeline, NVIDIA Isaac
Sim~\cite{isaacsim} renders the scene and integrates rigid-body physics, and a hydrodynamics module derived from
MarineGym~\cite{marinegym} applies buoyancy, added mass, linear and quadratic damping and
per-thruster forces following standard marine-craft models~\cite{fossen}. Instead of a
simplified controller, the vehicle's actual ArduSub firmware runs in software-in-the-loop
mode, in lock-step with the physics at 62.5\,Hz, with the parameter set copied from the real
vehicle. The recorder stores the same streams as on the real vehicle, plus the flight controller's
velocity estimate and, where available, the ground-truth poses of the robot and the camera.

\subsection{Calibration Against the Real Vehicle}
\label{sec:fidelity}
Running the real firmware fixes how a joystick input becomes thruster commands; the simulator
must also match how the vehicle responds to these commands and what the camera sees.

\emph{Dynamics.} Each axis has added mass and linear and quadratic damping~\cite{fossen},
thrust follows the manufacturer's T200 thrust table~\cite{t200}, and the vehicle is neutrally buoyant. We
start from published BlueROV2 Heavy coefficients~\cite{vonbenzon2022,orjales2025} and check
them against calibration runs on the real vehicle. For rotation, we use runs that drive one
axis at a time and runs in which the pilot gives a command and then releases the joystick, so
that the motion dies out on its own. For each axis, we fit the added mass and damping so that
the simulated angular rates match those measured by the IMU; because the fitted values differ
slightly between runs, we keep their range rather than a single value. Roll and pitch also
depend on how far the center of buoyancy sits above the center of gravity: when the recorded
commands are replayed in the simulator, the measured offset of 0.024\,m~\cite{orjales2025}
gives a smaller attitude error than the CAD value of 0.01\,m~\cite{vonbenzon2022} in all
three roll and pitch runs. For translation, a Doppler velocity log measures steady speed
reliably but is too noisy for accelerations, so we check the steady-state response, which
confirms the published coefficients~\cite{vonbenzon2022}; heave keeps the published values.

Fig.~\ref{fig:steady} compares the steady response of the real and the simulated vehicle for
every joystick input held in these runs. For deflections of 25--75\%, the median difference
is below 8\% on all three axes, and at high deflection both yaw rates level off at about
105--125$^\circ$/s. The steady state is what matters most for our data: real pilots hold a
non-zero surge or sway command constant for at least one second during about 67\% and 77\% of
the time, and never exceed half deflection. Because the calibration runs give a range rather
than a single value for each parameter, we define three dynamics levels for surge, roll,
pitch and yaw: the fitted center and two levels a third of the way from it toward each end of
the range, with damping and added mass moved together. Recordings use one of these three
levels, chosen at random for each recording and stored with it, so that the data cover the
spread measured on the real vehicle.

\emph{Appearance.} The water effect follows the revised underwater image formation
model~\cite{akkaynak2018}, $I=J\,e^{-\beta^D d}+B^\infty(1-e^{-\beta^B d})$, where $J$ is the
image without water and $d$ the distance. For the deck and diver scenes, we fit its parameters
to Blue Grotto footage; for the other scenes, which have no real counterpart, we fit them with
SUCRe~\cite{sucre} on public underwater scenes~\cite{seathru,seathrunerf,seasplat} and define
four water conditions, from clear to turbid. The water effect is added after rendering with
the simulator's exact depth, directly for the shipwreck, oyster-reef and structure scenes and
through the image decoder of WaterGen~\cite{watergen} for the deck and diver scenes.

%% file: sections/03_dataset.tex
\section{The \dataset{} Dataset}
\label{sec:dataset}

\begin{figure*}[t]
  \centering
  \includegraphics[width=\textwidth]{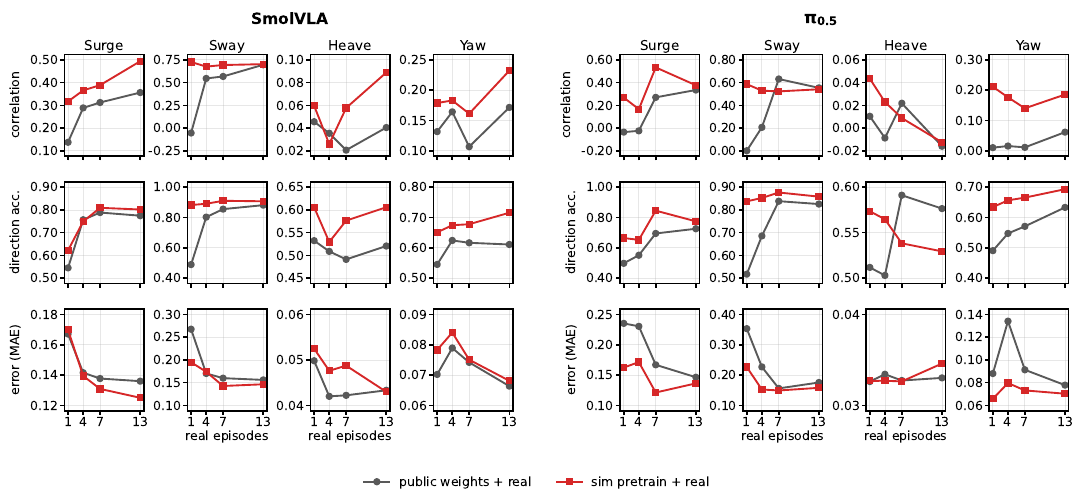}
  \caption{Fine-tuning from public weights versus fine-tuning after pretraining on simulated
  deck episodes, as a function of the real episodes used, for SmolVLA (left) and $\pi_{0.5}$
  (right). Rows: correlation with the pilot (pooled over test episodes), direction accuracy,
  error; columns: surge, sway, heave, yaw. }
  \label{fig:channels}
\end{figure*}

With this platform in place, pilots recorded two kinds of demonstrations: real ones at a
single dive site and simulated ones in the calibrated simulation pipeline of
Sec.~\ref{sec:platform}.
Together they form \dataset{}.

\subsection{Real-World Demonstrations}
Real-world data were recorded at Blue Grotto, Florida, a freshwater spring dive site with a
submerged deck, on six field days. Experienced pilots performed nine task types; the
two main ones, with at least 25 minutes each, are scanning the submerged deck and following a
diver. Each recording stores the forward camera stream, the IMU, pressure depth and the RC
channels. After manual cleaning, 2.0 hours form the action-training set
(Table~\ref{tab:stats}): 53 episodes with 71{,}741 frames. Deck scan is used in the sim-to-real experiment of Sec.~\ref{sec:exp_real}; the other real
tasks have too few episodes to benchmark and are released as part of the dataset.
A further 2.0 hours of real footage without a usable action target,
from sensor calibration, free driving and cave exploration, are
released for visual pretraining and domain statistics.

\subsection{Simulated Demonstrations}
Pilots drove the simulation pipeline of Sec.~\ref{sec:sim} through seven task families. They follow the
applications of Sec.~\ref{sec:intro} and cover the main ways a pilot moves relative to a
target: orbiting a structure (a shipwreck, an electrical box, or a
submerged deck), sweeping a wind-turbine foundation up
and down, following a subsea pipeline, searching for and inspecting oyster clusters on the
seabed, and following animated divers that swim long, irregular trajectories. Deck scan and
diver following are also the two main real tasks, so they link the simulated part to the
real one; the other scenes extend the data to sites that a real vehicle reaches only rarely
or at high cost. The simulated part contains 394 recordings and 17.9 hours (Table~\ref{tab:stats}); for training,
they are cut at the boundaries of their task segments into 513 task episodes.

\subsection{Language Instructions}
Each episode is labeled with one instruction that states the goal of its task. The deck,
electrical-box and shipwreck scans name the target and the sweep direction, as in ``Scan the
shipwreck from right to left''; the other families name only the task, as in ``Follow the
pipeline''. The simulated part uses 13 instructions in total.

\begin{table}[t]
  \centering
  \caption{\dataset{} statistics. Count: episodes (real) or recordings (simulated).}
  \label{tab:stats}
  \begin{tabular}{llrrr}
    \toprule
    Part & Task family & Count & Frames & Hours \\
    \midrule
    \multirow{3}{*}{Real} & Diver following  & 10  & 23{,}511  & 0.65 \\
                          & Deck scan        & 18  & 21{,}663  & 0.60 \\
                          & Other (7 tasks)  & 25  & 26{,}567  & 0.74 \\
    \midrule
    \multirow{7}{*}{Sim}  & Diver following  & 137 & 202{,}721 & 5.63 \\
                          & Shipwreck scan   & 79  & 169{,}609 & 4.71 \\
                          & Oyster-reef insp.& 78  & 120{,}408 & 3.34 \\
                          & Deck scan        & 53  & 77{,}703  & 2.16 \\
                          & Pipeline follow  & 23  & 44{,}589  & 1.24 \\
                          & Wind turbine     & 10  & 19{,}026  & 0.53 \\
                          & Electrical box   & 14  & 12{,}111  & 0.34 \\
    \midrule
    \multicolumn{2}{l}{Sim total}  & 394 & 646{,}167 & 17.95 \\
    \multicolumn{2}{l}{Real total} & 53  & 71{,}741  & 1.99 \\
    \bottomrule
  \end{tabular}
  \vspace{-0.3in}
\end{table}

%% file: sections/05_exp1_sim_value.tex
\section{Experiment 1: Sim-to-Real Transfer of the Dataset}
\label{sec:exp_real}
We first test whether the simulated part of \dataset{} helps a policy learn from limited
real data. To test it, we pretrain a policy on simulated
episodes of a task, fine-tune it on increasing amounts of real data of the same task, and
compare it with the same policy fine-tuned from its public weights on the same real data
(Fig.~\ref{fig:channels}). If the simulated part transfers, the pretrained policy should
score higher than the one from public weights on the same real data, and so reach a given
performance with less of it.

\subsection{Setup}
We run this test on deck scan, which is recorded both at Blue Grotto and in simulation. We use
two VLA models, the compact SmolVLA~\cite{smolvla} and the larger $\pi_{0.5}$~\cite{pi05}.
Both receive the forward image, the six-dimensional inertial state and the task instruction,
and predict the four joystick channels as absolute commands.

The sim-pretrained policy is first trained for 6{,}000 steps on 61 episodes (68{,}101 frames) cut from the simulated deck
recordings. The real-world training set has 13 deck episodes (16{,}849 frames, 28 minutes). We fine-tune each policy on nested subsets of 1, 4, 7 and all 13 of these
episodes, that is, 3.7, 8.8, 14.5 and 28 minutes of driving. Fine-tuning uses each
model's default learning rate, a batch of 64 for SmolVLA and 16 for $\pi_{0.5}$, at most
6{,}000 steps and early stopping on one fixed real validation episode outside these subsets;
for SmolVLA, the line from public weights is trained for a fixed 6{,}000 steps. All policies
are tested on the four deck episodes of one recording day (3{,}765 frames), which are used
neither for training nor for model selection.

For each channel we report three measures over the test frames: the correlation with the
pilot's commands, pooled over the four test episodes; the direction accuracy, the fraction of
frames with a clear pilot command ($|a|>0.05$) in which the prediction has the same sign, which
shows directly whether a scan sweeps the correct way; and the mean absolute error. The error
alone is not enough, because a policy that always outputs the average command can have a low
error without following the pilot at all; the correlation and the direction accuracy reveal
such a policy.

\subsection{Results}
Fig.~\ref{fig:channels} shows the three measures for every channel as the real data grow.
Pretraining helps on every channel the pilot actually uses in a deck scan. With a single real
episode (3.7 minutes), the policies fine-tuned from public weights do not learn the sweep: their
sway correlation is $-0.05$ for SmolVLA and 0.00 for $\pi_{0.5}$, and their sway direction is
right only about half of the time. Pretrained on our simulated deck episodes, the same policies
reach a sway correlation of 0.73 and 0.59 and a direction accuracy of 0.88 and 0.84 from the
same single episode. Surge and yaw follow the same pattern: from that episode, pretraining
raises their direction accuracy from 0.54 and 0.50 to 0.62 and 0.66 (surge) and from 0.55 and
0.49 to 0.65 and 0.63 (yaw), and it raises the correlation of both channels at every amount of
real data for both models. The error, averaged over the four channels, is also lower, by 0.015
and 0.055, with 95\% bootstrap intervals that exclude zero. Heave is the exception: in depth-hold
mode the flight controller keeps the depth, so the pilot gives a clear heave command in only
9\% of the test frames, against 24\% for yaw and over 88\% for surge and sway, and there is
little heave signal to learn from.

For SmolVLA, one real episode after pretraining matches the sway correlation and direction
accuracy that training from public weights reaches only with all 13 episodes (0.70 and
0.88), although its error is still higher (0.124 versus 0.100); pretraining thus takes the
place of about 24 minutes of real driving. As real data grow, the policies trained from
public weights catch up: their error is no longer significantly higher from 4 real episodes
on for SmolVLA and from 7 on for $\pi_{0.5}$. The simulated part is therefore most valuable
where real demonstrations are scarce, which is the situation it is built for.

%% file: sections/06_exp2_benchmark.tex
\section{Experiment 2: Benchmark on Simulated Data}
\label{sec:exp_sim}
With the simulated part shown to transfer to real data, we use the simulation pipeline to test how existing robot policies perform when they control the vehicle through joystick commands. We compare seven models from four families, trained on the same data, and
rank them by closed-loop success rate.

\subsection{Setup}
\emph{Data.} Training uses all seven simulated task families, with 13 instruction strings in total. Whole recording days go to the test set, because recordings from the same day share
lighting, water and pilot. Of the 513 simulated task episodes, 365 are used for training,
25 for validation and 123 for testing, with every task family in each; the validation set is used only for early stopping.

\emph{Models.} We train seven models from four families: imitation-learning policies without
language (a minimal convolutional behavior-cloning policy, BC, as well as ACT~\cite{act} and
Diffusion Policy~\cite{diffusionpolicy}); generalist VLA models (the compact
SmolVLA~\cite{smolvla} and the larger $\pi_{0.5}$~\cite{pi05}); VLA with memory
(MemoryVLA~\cite{memoryvla}, the only model here whose context reaches beyond the last two
frames); and underwater VLA (U0~\cite{usim_u0}, fine-tuned in simulation from GR00T~N1.5 and
the closest prior model to our setting). These models were built for other robots, whose
sensors and actions differ from ours, so we adapt their inputs and outputs and leave the rest
of each architecture unchanged. Each model receives the forward image and the
six-dimensional inertial state (MemoryVLA has no state input), which for U0 replaces its
original sensor readings, and predicts the four joystick channels as absolute commands with
its native action head, resized to these four channels for U0 and MemoryVLA. The VLA models
receive the task instruction; BC, ACT and Diffusion Policy, which have no language input,
receive a one-hot code of it appended to the state, and are trained from scratch apart from an ImageNet-pretrained ResNet-18 image
encoder~\cite{imagenet,resnet}; ACT, Diffusion Policy and SmolVLA use their LeRobot
implementations~\cite{lerobot}.

\emph{Training.} Each model follows its official fine-tuning recipe and learning rate and is
trained until its own loss on 2{,}000 fixed validation frames stops improving, checked every
32{,}000 training samples with a patience of five checks; the best checkpoint is kept. $\pi_{0.5}$ trains for at least
two passes before early stopping may apply, and MemoryVLA, fully fine-tuned with its original
real-robot recipe, was stopped after 8{,}900 steps, when its validation loss had reached a
plateau.  No hyperparameter is tuned for the underwater tasks, so the results show how each
model performs with its default recipe rather than the best it could reach. All results use
a single seed.

\subsection{Closed-Loop Evaluation}
Every trained model is deployed in the same simulation pipeline that the pilots used to
record the demonstrations (Sec.~\ref{sec:sim}): its predicted channels are sent to the
flight-controller firmware as RC overrides, at the recording pace, and the water is rendered with one of
the appearance settings used for the training frames of each family. The policy runs at
10\,Hz and re-plans after executing five steps (0.5\,s). The targets, the seabed and the
rocks are given collision geometry so that a policy cannot pass through them.

The benchmark covers six of the seven task families: diver following, deck scan,
electrical-box scan, shipwreck scan, oyster-reef inspection and pipeline following. The
wind-turbine scan is left out because it makes up less than 3\% of the training frames and
its demonstrations start far from any pose a policy could be placed at. Each family starts
from one fixed pose. Diver following and oyster inspection are run five times; the three
scans are run five times in each sweep direction and pipeline following five times from each
end of the pipe, so these four families are run ten times each. This gives 50 episodes per
model. An episode ends at a time limit set per task from the human demonstrations, three to
four times their mean duration for the scans.

Each episode is first judged automatically from the ground-truth poses logged at 10\,Hz. A
scan succeeds when every scored face of the target has been seen, within the sensing range of
that task, over at least 75\% of its surface points, and when the vehicle has turned at least
270$^\circ$ around the target in the instructed direction. The deck scan instead requires its three scored faces in the instructed order; only three
are scored because the vehicle can reach only three sides of the real deck at Blue Grotto.  Oyster inspection succeeds when at least 75\% of
the 2{,}183 oysters on the reef have been seen at close range, and pipeline following when at
least 75\% of the task segment has been seen and the vehicle has reached within 5\,m of its far
end. Diver following succeeds when the diver stays in the image for at
least 90\% of the time, the vehicle never comes closer than 1\,m, and the median distance over
the last 30\,s is at most 15\,m. Diver following is judged over the first loop of the diver's path, which matches the length
of the human demonstrations. Every verdict was then checked by a human reviewer on the recorded
video. Because a binary verdict hides how much of a task was done, each episode is also given
a completion value, the mean fraction of each requirement that was met, capped at one per
requirement.

\subsection{Results}
Table~\ref{tab:closedloop} summarizes the 350 episodes.

\begin{table}[t]
  \centering
  \caption{Closed-loop results in simulation, fixed start: number of episodes out of 50 per
  model. Full: all criteria met; Coll.-free: full and without any collision; $\geq$75/50/25:
  completion at least 75/50/25\%.}
  \label{tab:closedloop}
  \setlength{\tabcolsep}{3pt}
  \small
  \begin{tabular}{l cc ccc}
    \toprule
    Model & Full & Coll.-free & $\geq$75 & $\geq$50 & $\geq$25 \\
    \midrule
    BC               &  0/50 &  0/50 &  3/50 &  8/50 & 21/50 \\
    U0               &  0/50 &  0/50 &  4/50 & 11/50 & 24/50 \\
    SmolVLA          &  2/50 &  0/50 &  3/50 &  9/50 & 21/50 \\
    MemoryVLA        &  2/50 &  2/50 &  5/50 &  8/50 & 14/50 \\
    ACT              &  9/50 &  3/50 & 10/50 & 15/50 & 26/50 \\
    Diff.\ Policy    &  9/50 &  6/50 & 14/50 & 22/50 & 24/50 \\
    $\pi_{0.5}$      & 10/50 &  2/50 & 14/50 & 20/50 & 25/50 \\
    \bottomrule
  \end{tabular}
  \vspace{-0.15in}
\end{table}

No policy completes a task family reliably, with a single exception. Under the full criterion, the best
policies succeed in only 18--20\% of their episodes ($\pi_{0.5}$, ACT and Diffusion Policy).
Per task family, the best rates are 50\% on the shipwreck ($\pi_{0.5}$), 30\% on the deck (ACT
and Diffusion Policy) and 20\% on the electrical box ($\pi_{0.5}$). Requiring the episode to be
free of collisions lowers the overall rates to 4--12\%, because 72\%
of all episodes contain at least one collision. The one exception is diver following by
Diffusion Policy, which succeeds in all five episodes and keeps the diver in the image for a
median of 148\,s of the 150\,s loop.

These rankings do not follow model size, and pretraining on land data gives no advantage. ACT
and Diffusion Policy, which have no pretrained vision-language backbone and receive the
instruction only as a one-hot index, match $\pi_{0.5}$ under the full criterion. The other
three vision-language-action models fall well below it: SmolVLA succeeds in 4\% of its
episodes, MemoryVLA in 4\%, and U0 never succeeds, which places U0 with the BC baseline.
MemoryVLA, the only model that sees beyond the last two frames, thus gains nothing from its
memory, and U0, the one VLA built for underwater vehicles, gains nothing from its underwater
pretraining, which targets thruster commands rather than the joystick.

Where the policies succeed, they do so mostly in one sweep direction. $\pi_{0.5}$ completes the
shipwreck scan in all five right-to-left episodes but in none of the left-to-right ones, and
across all models the deck scan succeeds in 8 left-to-right but only 2 right-to-left episodes.
The two tasks that require the longest horizons almost always fail: pipeline following never
succeeds in 70 episodes and reaches the far end only twice, and oyster inspection never
succeeds, with the best episode seeing 37\% of the oysters, because the policies that find the
reef stay at the first patch they reach instead of moving on to the others. The completion
levels separate two ways of falling short: BC and U0 reach 25\% completion on the electrical box
in every episode but never complete it, covering neither every face nor the full angle, whereas $\pi_{0.5}$ reaches 75\%
completion on the shipwreck in 70\% of its episodes and succeeds in half.

\subsection{Failure Analysis}
The completion levels show where the episodes break down. Relaxing the requirement to 25\%
completion raises every model to 28--52\%, so most policies start the task correctly by
approaching the target and beginning to orbit or follow it. Fewer than half of the episodes,
however, reach 50\% completion for any model, which means the failures lie in sustaining
the task rather than in starting it, for large and small models alike. The trajectories point to
four recurring causes: drifting away from the target (103 of 350
episodes), contact with the seabed (139), spinning in place (58), and stopping an orbit after
90--180$^\circ$ (46). These labels, computed from the logged trajectories, differ by model more than by task: $\pi_{0.5}$ reaches the target and then spins beside it, ACT either
completes the orbit or drifts off and stops commanding, Diffusion Policy runs into the
structure and stays there, and MemoryVLA touches the seabed in most episodes and collides
most often. For diver following, the policies other than Diffusion Policy stay within 15\,m
of the diver but do not keep the camera on it, so the failure is one of heading rather than
of speed.

%% file: sections/08_conclusion.tex
\section{Conclusion}
\label{sec:conclusion}
We proposed to learn underwater VLA policies from the joystick commands that pilots already
send in every teleoperated dive. To this end, we presented \dataset{}, whose real and simulated
demonstrations are both recorded by human pilots through the control chain of a real BlueROV2,
together with a closed-loop benchmark in simulation.

Our experiments lead to two findings. First, the simulated demonstrations help a policy learn
from real data: after pretraining on them, a policy learns the deck scan from a single real
episode about as well as it otherwise does from all 13. Second, current policies remain far
from solving these tasks. Trained with their default recipes, even the best of them succeed
in only 18--20\% of the closed-loop episodes, and none completes the long pipeline and reef
tasks. Because large pretrained VLA models do no better than small policies trained from
scratch, and memory does not help either, the gap lies less in model size than in keeping a
task going over a long time, and it shows how far land-trained models still are from working
underwater. \dataset{} makes this gap measurable, and its simulation pipeline lets pilots
record more demonstrations of exactly these long tasks.

In future work, we will test the policies on the real vehicle, extend the data to more sites,
and identify which parts of the simulation pipeline matter most for transfer.

%% file: refs.bib
@IEEEtranBSTCTL{IEEEexample:BSTcontrol,
  CTLuse_forced_etal       = "yes",
  CTLmax_names_forced_etal = "3",
  CTLnames_show_etal       = "1",
  CTLdash_repeated_names   = "no" }

@inproceedings{rt2,
  author    = {Brianna Zitkovich and Tianhe Yu and Sichun Xu and Peng Xu and Ted Xiao and Fei Xia and Jialin Wu and Paul Wohlhart and others},
  title     = {{RT-2}: Vision-Language-Action Models Transfer Web Knowledge to Robotic Control},
  booktitle = {Conference on Robot Learning (CoRL)},
  series    = {Proceedings of Machine Learning Research},
  volume    = {229},
  pages     = {2165--2183},
  publisher = {PMLR},
  year      = {2023}
}

@inproceedings{openvla,
  author    = {Moo Jin Kim and Karl Pertsch and Siddharth Karamcheti and Ted Xiao and Ashwin Balakrishna and Suraj Nair and Rafael Rafailov and Ethan P. Foster and Pannag R. Sanketi and Quan Vuong and Thomas Kollar and Benjamin Burchfiel and Russ Tedrake and Dorsa Sadigh and Sergey Levine and Percy Liang and Chelsea Finn},
  title     = {{OpenVLA}: An Open-Source Vision-Language-Action Model},
  booktitle = {Conference on Robot Learning (CoRL)},
  series    = {Proceedings of Machine Learning Research},
  volume    = {270},
  pages     = {2679--2713},
  publisher = {PMLR},
  year      = {2025}
}

@inproceedings{octo,
  author    = {Dibya Ghosh and Homer Rich Walke and Karl Pertsch and Kevin Black and Oier Mees and Sudeep Dasari and Joey Hejna and Tobias Kreiman and Charles Xu and Jianlan Luo and You Liang Tan and Lawrence Yunliang Chen and Quan Vuong and Ted Xiao and Pannag R. Sanketi and Dorsa Sadigh and Chelsea Finn and Sergey Levine},
  title     = {{Octo}: An Open-Source Generalist Robot Policy},
  booktitle = {Robotics: Science and Systems (RSS)},
  year      = {2024},
  doi       = {10.15607/RSS.2024.XX.090}
}

@inproceedings{pi0,
  author    = {Kevin Black and Noah Brown and Danny Driess and Adnan Esmail and Michael Robert Equi and Chelsea Finn and Niccolo Fusai and Lachy Groom and Karol Hausman and Brian Ichter and Szymon Jakubczak and Tim Jones and Liyiming Ke and Sergey Levine and Adrian Li-Bell and Mohith Mothukuri and Suraj Nair and Karl Pertsch and Lucy Xiaoyang Shi and Laura Smith and James Tanner and Quan Vuong and Anna Walling and Haohuan Wang and Ury Zhilinsky},
  title     = {$\pi_0$: A Vision-Language-Action Flow Model for General Robot Control},
  booktitle = {Robotics: Science and Systems (RSS)},
  year      = {2025},
  doi       = {10.15607/RSS.2025.XXI.010}
}

@inproceedings{pi05,
  author    = {Kevin Black and Noah Brown and James Darpinian and Karan Dhabalia and Danny Driess and Adnan Esmail and Michael Robert Equi and Chelsea Finn and others},
  title     = {$\pi_{0.5}$: A Vision-Language-Action Model with Open-World Generalization},
  booktitle = {Conference on Robot Learning (CoRL)},
  series    = {Proceedings of Machine Learning Research},
  volume    = {305},
  pages     = {17--40},
  publisher = {PMLR},
  year      = {2025}
}

@article{gr00t,
  author    = {{NVIDIA} and Johan Bjorck and Fernando Casta{\~n}eda and Nikita Cherniadev and Xingye Da and Runyu Ding and Linxi Fan and Yu Fang and others},
  title     = {{GR00T} {N1}: An Open Foundation Model for Generalist Humanoid Robots},
  journal   = {arXiv preprint arXiv:2503.14734},
  year      = {2025}
}

@article{smolvla,
  author    = {Mustafa Shukor and Dana Aubakirova and Francesco Capuano and Pepijn Kooijmans and Steven Palma and Adil Zouitine and Michel Aractingi and Caroline Pascal and Martino Russi and Andres Marafioti and Simon Alibert and Matthieu Cord and Thomas Wolf and Remi Cadene},
  title     = {{SmolVLA}: A Vision-Language-Action Model for Affordable and Efficient Robotics},
  journal   = {arXiv preprint arXiv:2506.01844},
  year      = {2025}
}

@inproceedings{oxe,
  author    = {{Open X-Embodiment Collaboration} and Abby O'Neill and Abdul Rehman and Abhiram Maddukuri and Abhishek Gupta and Abhishek Padalkar and Abraham Lee and others},
  title     = {{Open X-Embodiment}: Robotic Learning Datasets and {RT-X} Models},
  booktitle = {IEEE International Conference on Robotics and Automation (ICRA)},
  pages     = {6892--6903},
  year      = {2024},
  doi       = {10.1109/ICRA57147.2024.10611477}
}

@inproceedings{bridgev2,
  author    = {Homer Rich Walke and Kevin Black and Tony Z. Zhao and Quan Vuong and Chongyi Zheng and Philippe Hansen-Estruch and Andre Wang He and Vivek Myers and Moo Jin Kim and Max Du and Abraham Lee and Kuan Fang and Chelsea Finn and Sergey Levine},
  title     = {{BridgeData} {V2}: A Dataset for Robot Learning at Scale},
  booktitle = {Conference on Robot Learning (CoRL)},
  series    = {Proceedings of Machine Learning Research},
  volume    = {229},
  pages     = {1723--1736},
  publisher = {PMLR},
  year      = {2023}
}

@inproceedings{act,
  author    = {Tony Z. Zhao and Vikash Kumar and Sergey Levine and Chelsea Finn},
  title     = {Learning Fine-Grained Bimanual Manipulation with Low-Cost Hardware},
  booktitle = {Robotics: Science and Systems (RSS)},
  year      = {2023},
  doi       = {10.15607/RSS.2023.XIX.016}
}

@inproceedings{diffusionpolicy,
  author    = {Cheng Chi and Siyuan Feng and Yilun Du and Zhenjia Xu and Eric Cousineau and Benjamin Burchfiel and Shuran Song},
  title     = {Diffusion Policy: Visuomotor Policy Learning via Action Diffusion},
  booktitle = {Robotics: Science and Systems (RSS)},
  year      = {2023},
  doi       = {10.15607/RSS.2023.XIX.026}
}

@misc{lerobot,
  author       = {Cadene, Remi and Alibert, Simon and Soare, Alexander and Gallouedec, Quentin and Zouitine, Adil and Palma, Steven and Kooijmans, Pepijn and Aractingi, Michel and Shukor, Mustafa and Aubakirova, Dana and Russi, Martino and Capuano, Francesco and Pascal, Caroline and Choghari, Jade and Meftah, Khalil and Ellerbach, Maxime and Moss, Jess and Wolf, Thomas},
  title        = {{LeRobot}: State-of-the-art Machine Learning for Real-World Robotics in {PyTorch}},
  howpublished = {\url{https://github.com/huggingface/lerobot}},
  year         = {2024}
}

@inproceedings{omninav,
  author    = {Xinda Xue and Junjun Hu and Minghua Luo and Shichao Xie and Jintao Chen and Zixun Xie and Kuichen Quan and Wei Guo and Zedong Chu and Mu Xu and Zhengzhou Zhu},
  title     = {{OmniNav}: A Unified Framework for Prospective Exploration and Visual-Language Navigation},
  booktitle = {International Conference on Learning Representations (ICLR)},
  year      = {2026}
}

@article{usim_u0,
  author    = {Junwen Gu and Zhiheng Wu and Pengxuan Si and Shuang Qiu and Zhentao Zhang and Yukai Feng and Luoyang Sun and Laien Luo and Lianyi Yu and Jian Wang and Zhengxing Wu},
  title     = {{USIM} and {U0}: A Vision-Language-Action Dataset and Model for General Underwater Robots},
  journal   = {arXiv preprint arXiv:2510.07869},
  year      = {2025}
}

@inproceedings{navid,
  author    = {Jiazhao Zhang and Kunyu Wang and Rongtao Xu and Gengze Zhou and Yicong Hong and Xiaomeng Fang and Qi Wu and Zhizheng Zhang and He Wang},
  title     = {{NaVid}: Video-based {VLM} Plans the Next Step for Vision-and-Language Navigation},
  booktitle = {Robotics: Science and Systems (RSS)},
  year      = {2024},
  doi       = {10.15607/RSS.2024.XX.079}
}

@inproceedings{uninavid,
  author    = {Jiazhao Zhang and Kunyu Wang and Shaoan Wang and Minghan Li and Haoran Liu and Songlin Wei and Zhongyuan Wang and Zhizheng Zhang and He Wang},
  title     = {{Uni-NaVid}: A Video-based Vision-Language-Action Model for Unifying Embodied Navigation Tasks},
  booktitle = {Robotics: Science and Systems (RSS)},
  year      = {2025},
  doi       = {10.15607/RSS.2025.XXI.013}
}

@inproceedings{marinegym,
  author    = {Shuguang Chu and Zebin Huang and Yutong Li and Mingwei Lin and Dejun Li and Ignacio Carlucho and Yvan R. Petillot and Canjun Yang},
  title     = {{MarineGym}: A High-Performance Reinforcement Learning Platform for Underwater Robotics},
  booktitle = {IEEE/RSJ International Conference on Intelligent Robots and Systems (IROS)},
  pages     = {17146--17153},
  year      = {2025},
  doi       = {10.1109/IROS60139.2025.11246146}
}

@misc{isaacsim,
  author       = {{NVIDIA}},
  title        = {{Isaac Sim}: Robotics Simulation and Synthetic Data Generation},
  howpublished = {\url{https://developer.nvidia.com/isaac/sim}},
  note         = {Accessed: 2026-10-02}
}

@misc{ardusub,
  author       = {{ArduPilot Development Team}},
  title        = {{ArduSub}: Open-Source Autopilot Firmware for Underwater Vehicles ({ArduPilot} project)},
  howpublished = {\url{https://www.ardusub.com/} and \url{https://github.com/ArduPilot/ardupilot}},
  note         = {Accessed: 2026-10-02}
}

@book{fossen,
  author    = {Thor I. Fossen},
  title     = {Handbook of Marine Craft Hydrodynamics and Motion Control},
  edition   = {2nd},
  publisher = {John Wiley \& Sons},
  address   = {Hoboken, NJ, USA},
  year      = {2021},
  doi       = {10.1002/9781119575016}
}

@inproceedings{seathru,
  author    = {Derya Akkaynak and Tali Treibitz},
  title     = {{Sea-Thru}: A Method for Removing Water From Underwater Images},
  booktitle = {IEEE/CVF Conference on Computer Vision and Pattern Recognition (CVPR)},
  pages     = {1682--1691},
  year      = {2019},
  doi       = {10.1109/CVPR.2019.00178}
}

@inproceedings{akkaynak2018,
  author    = {Derya Akkaynak and Tali Treibitz},
  title     = {A Revised Underwater Image Formation Model},
  booktitle = {IEEE/CVF Conference on Computer Vision and Pattern Recognition (CVPR)},
  pages     = {6723--6732},
  year      = {2018},
  doi       = {10.1109/CVPR.2018.00703}
}

@article{auv_inspection_review,
  author  = {Franka Nauert and Peter Kampmann},
  title   = {Inspection and Maintenance of Industrial Infrastructure with Autonomous Underwater Robots},
  journal = {Frontiers in Robotics and AI},
  volume  = {10},
  pages   = {1240276},
  year    = {2023},
  doi     = {10.3389/frobt.2023.1240276}
}

@article{robot_vla_survey,
  author  = {Ranjan Sapkota and Yang Cao and Konstantinos I. Roumeliotis and Manoj Karkee},
  title   = {Vision-Language-Action ({VLA}) Models: Concepts, Progress, Applications and Challenges},
  journal = {arXiv preprint arXiv:2505.04769},
  year    = {2025}
}

@inproceedings{uw_aquabot,
  author    = {Ruoshi Liu and Huy Ha and Mengxue Hou and Shuran Song and Carl Vondrick},
  title     = {Self-Improving Autonomous Underwater Manipulation},
  booktitle = {IEEE International Conference on Robotics and Automation (ICRA)},
  pages     = {16915--16922},
  year      = {2025},
  doi       = {10.1109/ICRA55743.2025.11128759}
}

@article{uw_rovtether_scidata,
  author  = {Juri Khanmeh and Bilal Wehbe and Enrico Simetti and Giovanni Indiveri},
  title   = {Tracking Data of a Remotely Operated Vehicle and its Tether Using a Motion Capture System and a Tension Sensor},
  journal = {Scientific Data},
  volume  = {13},
  number  = {1},
  pages   = {35},
  year    = {2026},
  doi     = {10.1038/s41597-025-06347-0}
}

@inproceedings{uw_nav2goal,
  author    = {Travis Manderson and Gamboa Higuera, Juan Camilo and Stefan Wapnick and Jean-Fran{\c{c}}ois Tremblay and Florian Shkurti and David Meger and Gregory Dudek},
  title     = {Vision-Based Goal-Conditioned Policies for Underwater Navigation in the Presence of Obstacles},
  booktitle = {Robotics: Science and Systems (RSS)},
  year      = {2020},
  doi       = {10.15607/RSS.2020.XVI.048}
}

@article{uw_uloha,
  author  = {Masato Kobayashi and Takeru Tsunoori},
  title   = {{ULOHA}: An Underwater Bimanual Robot System for Robot Learning},
  journal = {arXiv preprint arXiv:2609.19200},
  year    = {2026}
}

@article{uw_biaqua,
  author  = {Takeru Tsunoori and Masato Kobayashi and Yuki Uranishi},
  title   = {{Bi-AQUA}: Bilateral Control-Based Imitation Learning for Underwater Robot Arms via Lighting-Aware Action Chunking with Transformers},
  journal = {arXiv preprint arXiv:2511.16050},
  year    = {2025}
}

@inproceedings{uw_uivnav,
  author    = {Xiaomin Lin and Nare Karapetyan and Kaustubh Joshi and Tianchen Liu and Nikhil Chopra and Miao Yu and Pratap Tokekar and Yiannis Aloimonos},
  title     = {{UIVNAV}: Underwater Information-driven Vision-based Navigation via Imitation Learning},
  booktitle = {IEEE International Conference on Robotics and Automation (ICRA)},
  pages     = {5250--5256},
  year      = {2024},
  doi       = {10.1109/ICRA57147.2024.10611203}
}

@article{offshore_inspection_review,
  author  = {Maricruz Fun Sang Cepeda and Marcos de Souza Freitas Machado and Fabr{\'i}cio Hudson Sousa Barbosa and Douglas Santana Souza Moreira and Maria Jos{\'e} Legaz Almansa and Marcelo Igor Louren{\c{c}}o de Souza and Jean-David Caprace},
  title   = {Exploring Autonomous and Remotely Operated Vehicles in Offshore Structure Inspections},
  journal = {Journal of Marine Science and Engineering},
  volume  = {11},
  number  = {11},
  pages   = {2172},
  year    = {2023},
  doi     = {10.3390/jmse11112172}
}

@article{netpen_visual_servo,
  author  = {Waseem Akram and Alessandro Casavola and Nadir Kapetanovi{\'c} and Nikola Mi{\v{s}}kovi{\'c}},
  title   = {A Visual Servoing Scheme for Autonomous Aquaculture Net Pens Inspection Using {ROV}},
  journal = {Sensors},
  volume  = {22},
  number  = {9},
  pages   = {3525},
  year    = {2022},
  doi     = {10.3390/s22093525}
}

@inproceedings{oystersim,
  author    = {Xiaomin Lin and Nitesh Jha and Mayank Joshi and Nare Karapetyan and Yiannis Aloimonos and Miao Yu},
  title     = {{OysterSim}: Underwater Simulation for Enhancing Oyster Reef Monitoring},
  booktitle = {OCEANS 2022, Hampton Roads},
  pages     = {1--6},
  year      = {2022},
  doi       = {10.1109/OCEANS47191.2022.9977233}
}

@inproceedings{dream_uwmonitor,
  author    = {Zhenqi Wu and Abhinav Modi and Angelos Mavrogiannis and Kaustubh Joshi and Nikhil Chopra and Yiannis Aloimonos and Nare Karapetyan and Ioannis Rekleitis and Xiaomin Lin},
  title     = {{DREAM}: Domain-aware Reasoning for Efficient Autonomous Underwater Monitoring},
  booktitle = {IEEE International Conference on Robotics and Automation (ICRA)},
  pages     = {17856--17863},
  year      = {2026},
  doi       = {10.1109/ICRA57385.2026.11697042}
}

@article{caddy,
  author  = {Nikola Mi{\v{s}}kovi{\'c} and Marco Bibuli and Andreas Birk and Massimo Caccia and Murat Egi and Karl Grammer and Alessandro Marroni and Jeff Neasham and Antonio Pascoal and Antonio Vasilijevi{\'c} and Zoran Vuki{\'c}},
  title   = {{CADDY}---Cognitive Autonomous Diving Buddy: Two Years of Underwater Human-Robot Interaction},
  journal = {Marine Technology Society Journal},
  volume  = {50},
  number  = {4},
  pages   = {54--66},
  year    = {2016},
  doi     = {10.4031/MTSJ.50.4.11}
}

@misc{osha_diving,
  author       = {{Occupational Safety and Health Administration (OSHA)}},
  title        = {Commercial Diving Operations, 29 {CFR} 1910 Subpart {T}: {\S}1910.424 {SCUBA} Diving and {\S}1910.425 Surface-Supplied Air Diving},
  howpublished = {\url{https://www.osha.gov/laws-regs/regulations/standardnumber/1910/1910.424}; \url{https://www.osha.gov/laws-regs/regulations/standardnumber/1910/1910.425}},
  year         = {2026},
  note         = {Accessed: 2026-10-04}
}

@article{petillot2019,
  author  = {Yvan R. Petillot and Gianluca Antonelli and Giuseppe Casalino and Fausto Ferreira},
  title   = {Underwater Robots: From Remotely Operated Vehicles to Intervention-Autonomous Underwater Vehicles},
  journal = {IEEE Robotics \& Automation Magazine},
  volume  = {26},
  number  = {2},
  pages   = {94--101},
  year    = {2019},
  doi     = {10.1109/MRA.2019.2908063}
}

@article{sharc,
  author  = {Amy Phung and Gideon Billings and Andrea F. Daniele and Matthew R. Walter and Richard Camilli},
  title   = {Enhancing Scientific Exploration of the Deep Sea through Shared Autonomy in Remote Manipulation},
  journal = {Science Robotics},
  volume  = {8},
  number  = {81},
  pages   = {eadi5227},
  year    = {2023},
  doi     = {10.1126/scirobotics.adi5227}
}

@article{subsea_hmi_review,
  author  = {Adnan Abdullah and David Blow and Ruo Chen and Thanakon Uthai and Eric Jing Du and Md Jahidul Islam},
  title   = {Human-Machine Interfaces for Subsea Telerobotics: From Soda-straw to Natural Language Interactions},
  journal = {arXiv preprint arXiv:2412.01753},
  year    = {2024}
}

@inproceedings{ridao_iauv,
  author    = {Pere Ridao and Marc Carreras and David Ribas and Pedro J. Sanz and Gabriel Oliver},
  title     = {Intervention {AUVs}: The Next Challenge},
  booktitle = {19th IFAC World Congress},
  series    = {IFAC Proceedings Volumes},
  volume    = {47},
  number    = {3},
  pages     = {12146--12159},
  year      = {2014},
  doi       = {10.3182/20140824-6-ZA-1003.02819}
}

@article{hover2012hull,
  author  = {Franz S. Hover and Ryan M. Eustice and Ayoung Kim and Brendan Englot and Hordur Johannsson and Michael Kaess and John J. Leonard},
  title   = {Advanced Perception, Navigation and Planning for Autonomous In-Water Ship Hull Inspection},
  journal = {The International Journal of Robotics Research},
  volume  = {31},
  number  = {12},
  pages   = {1445--1464},
  year    = {2012},
  doi     = {10.1177/0278364912461059}
}

@article{xiang2016cable,
  author  = {Xianbo Xiang and Caoyang Yu and Zemin Niu and Qin Zhang},
  title   = {Subsea Cable Tracking by Autonomous Underwater Vehicle with Magnetic Sensing Guidance},
  journal = {Sensors},
  volume  = {16},
  number  = {8},
  pages   = {1335},
  year    = {2016},
  doi     = {10.3390/s16081335}
}

@article{uw_ai_review,
  author  = {Leif Christensen and Jos{\'e} de Gea Fern{\'a}ndez and Marc Hildebrandt and Christian Ernst Siegfried Koch and Bilal Wehbe},
  title   = {Recent Advances in {AI} for Navigation and Control of Underwater Robots},
  journal = {Current Robotics Reports},
  volume  = {3},
  number  = {4},
  pages   = {165--175},
  year    = {2022},
  doi     = {10.1007/s43154-022-00088-3}
}

@article{uw_full_autonomy_review,
  author  = {Ali Rohan and Hamidreza Farhadi Tolie and Md Junayed Hasan and Somasundar Kannan},
  title   = {Full Autonomy in Underwater Robotics Systems: A Realistic Prospect?},
  journal = {Engineering Applications of Artificial Intelligence},
  volume  = {162},
  pages   = {112638},
  year    = {2025},
  doi     = {10.1016/j.engappai.2025.112638}
}

@article{vonbenzon2022,
  author  = {von Benzon, Malte and Sørensen, Fredrik Fogh and Uth, Esben and Jouffroy, Jerome and Liniger, Jesper and Pedersen, Simon},
  title   = {An Open-Source Benchmark Simulator: Control of a {BlueROV2} Underwater Robot},
  journal = {Journal of Marine Science and Engineering},
  volume  = {10},
  number  = {12},
  pages   = {1898},
  year    = {2022},
  doi     = {10.3390/jmse10121898}
}

@article{orjales2025,
  author  = {F{\'e}lix Orjales and Juli{\'a}n Rodr{\'i}guez-Cortegoso and Enrique Fern{\'a}ndez-P{\'e}rez and Alejandro Romero and Vicente Diaz-Casas},
  title   = {Towards a Digital Twin for Open-Frame Underwater Vehicles Using Evolutionary Algorithms},
  journal = {Applied Sciences},
  volume  = {15},
  number  = {13},
  pages   = {7085},
  year    = {2025},
  doi     = {10.3390/app15137085}
}

@inproceedings{watergen,
  author    = {Jiayi Wu and Tianfu Wang and Tianyi Xiong and Dehao Yuan and Xiaomin Lin and Md Jahidul Islam and Cornelia Fermuller and Christopher Metzler and Yiannis Aloimonos},
  title     = {{WaterGen}: Decoupling Scene and Medium in Underwater Image Generation},
  booktitle = {European Conference on Computer Vision (ECCV)},
  series    = {Lecture Notes in Computer Science},
  publisher = {Springer},
  pages     = {151--170},
  year      = {2026},
  doi       = {10.1007/978-3-032-37189-8_9}
}

@inproceedings{sucre,
  author    = {Cl{\'e}mentin Boittiaux and Ricard Marxer and Claire Dune and Aur{\'e}lien Arnaubec and Maxime Ferrera and Vincent Hugel},
  title     = {{SUCRe}: Leveraging Scene Structure for Underwater Color Restoration},
  booktitle = {International Conference on 3D Vision (3DV)},
  pages     = {1488--1497},
  year      = {2024},
  doi       = {10.1109/3DV62453.2024.00148}
}

@inproceedings{seathrunerf,
  author    = {Deborah Levy and Amit Peleg and Naama Pearl and Dan Rosenbaum and Derya Akkaynak and Simon Korman and Tali Treibitz},
  title     = {{SeaThru-NeRF}: Neural Radiance Fields in Scattering Media},
  booktitle = {IEEE/CVF Conference on Computer Vision and Pattern Recognition (CVPR)},
  pages     = {56--65},
  year      = {2023},
  doi       = {10.1109/CVPR52729.2023.00014}
}

@inproceedings{seasplat,
  author    = {Daniel Yang and John J. Leonard and Yogesh Girdhar},
  title     = {{SeaSplat}: Representing Underwater Scenes with {3D} {Gaussian} Splatting and a Physically Grounded Image Formation Model},
  booktitle = {IEEE International Conference on Robotics and Automation (ICRA)},
  pages     = {7632--7638},
  year      = {2025},
  doi       = {10.1109/ICRA55743.2025.11128502}
}

@inproceedings{memoryvla,
  author    = {Hao Shi and Bin Xie and Yingfei Liu and Lin Sun and Fengrong Liu and Tiancai Wang and Erjin Zhou and Haoqiang Fan and Xiangyu Zhang and Gao Huang},
  title     = {{MemoryVLA}: Perceptual-Cognitive Memory in Vision-Language-Action Models for Robotic Manipulation},
  booktitle = {International Conference on Learning Representations (ICLR)},
  year      = {2026}
}

@misc{qgroundcontrol,
  author       = {{Dronecode Project}},
  title        = {{QGroundControl}: Ground Control Station},
  howpublished = {\url{https://qgroundcontrol.com/}},
  note         = {Accessed: 2026-10-18}
}

@misc{t200,
  author       = {{Blue Robotics}},
  title        = {{T200} Thruster: Technical Details and Performance Charts},
  howpublished = {\url{https://bluerobotics.com/store/thrusters/t100-t200-thrusters/t200-thruster-r2-rp/}},
  note         = {Accessed: 2026-10-18}
}

@inproceedings{imagenet,
  author    = {Jia Deng and Wei Dong and Richard Socher and Li-Jia Li and Kai Li and Li Fei-Fei},
  title     = {{ImageNet}: A Large-Scale Hierarchical Image Database},
  booktitle = {IEEE Conference on Computer Vision and Pattern Recognition (CVPR)},
  pages     = {248--255},
  year      = {2009},
  doi       = {10.1109/CVPR.2009.5206848}
}

@inproceedings{resnet,
  author    = {Kaiming He and Xiangyu Zhang and Shaoqing Ren and Jian Sun},
  title     = {Deep Residual Learning for Image Recognition},
  booktitle = {IEEE Conference on Computer Vision and Pattern Recognition (CVPR)},
  pages     = {770--778},
  year      = {2016},
  doi       = {10.1109/CVPR.2016.90}
}
